\documentclass[11pt,a4paper]{article}

\usepackage{times}
\usepackage{graphicx} 
\usepackage{sidecap}
\usepackage{subfigure}
\usepackage[small]{caption}

\usepackage{multicol}

\usepackage{latexsym}

\usepackage{appendix}
\usepackage{verbatim}

\usepackage{url}

\usepackage[usenames,dvipsnames,svgnames,table]{xcolor}  
\usepackage{todonotes} 
\usepackage[hyperref]{naaclhlt2018}
\aclfinalcopy 

\usepackage{amsmath}
\usepackage{amssymb}
\usepackage{mathtools, cuted} 
\usepackage{tabularx, booktabs}
\newcolumntype{C}{>{\centering\arraybackslash}X}
\newcolumntype{R}{>{\raggedleft\arraybackslash}X}
\newcolumntype{S}{>{\raggedleft\arraybackslash\hsize=.5\hsize}X}
\usepackage{latexsym}
\usepackage{url}
\usepackage{xspace}
\usepackage{bm,array}
\usepackage{amsfonts}
\usepackage{enumitem}
\usepackage{mathtools}
\usepackage{bm}
\usepackage{esvect}

\usepackage{cleveref}
\crefname{section}{§}{§§}
\Crefname{section}{§}{§§}
\Crefname{figure}{Fig.}{}
\Crefname{algorithm}{Algorithm}{}
\Crefname{equation}{Equation}{}

\usepackage{graphicx}
\usepackage{caption}

\definecolor{pred}{RGB}{1,0,254}
\definecolor{arg}{RGB}{133,16,193}

\usepackage{CJKutf8}

\renewcommand{\vec}[1]{{\boldsymbol{\mathbf{#1}}}}  

\newcommand{\Real}{\mathbb{R}}

\usepackage{booktabs}
\usepackage{multirow}

\title{{\em Halo}: Learning Semantics-Aware Representations for Cross-Lingual Information Extraction}

\author{Hongyuan Mei\thanks{\ \ equal contribution},\ \ Sheng Zhang\footnotemark[1],\ \  Kevin Duh,\ \  Benjamin Van Durme\\
  Center for Language and Speech Processing, Johns Hopkins University \\
  {\tt \{hmei,s.zhang,kevinduh,vandurme\}@cs.jhu.edu} 
  }
\date{}

\begin{document}
\maketitle
\begin{abstract}
Cross-lingual information extraction (CLIE) is an important and challenging task, especially in low resource scenarios. To tackle this challenge, we propose a training method, called {\em Halo}, which enforces the local region of each hidden state of a neural model to only generate target tokens with the same semantic structure tag. This simple but powerful technique enables a neural model to learn semantics-aware representations that are robust to noise, without introducing any extra parameter, thus yielding better generalization in both high and low resource settings.
\end{abstract}

\section{Introduction}
\begin{figure}[t]
\centering
\includegraphics[width=0.48\textwidth]{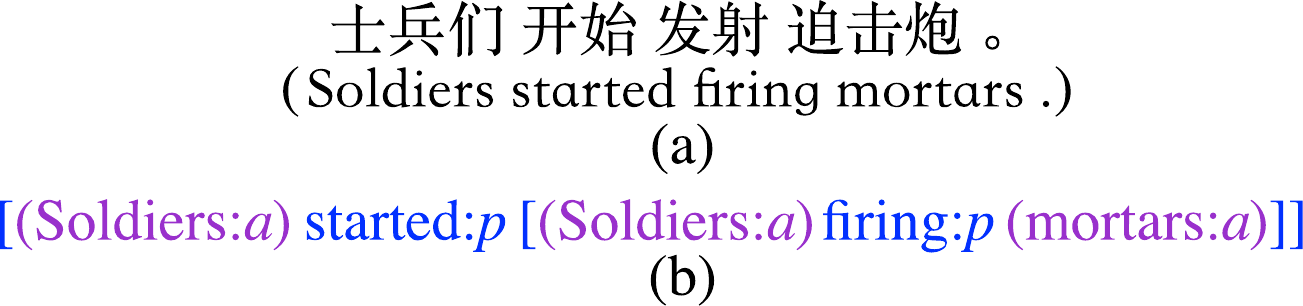}
\vspace{-20pt}
\caption{Example of cross-lingual information extraction: Chinese input text (a) and linearized English PredPatt output (b), where `:p' and \textcolor{blue}{blue} stand for predicate while `:a' and \textcolor{purple}{purple} denote argument.\label{fig:repr}}
\end{figure}

Cross-lingual information extraction (CLIE) is the task of distilling and representing factual information in a target language from the textual input in a source language~\cite{sudo2004cross,zhang-duh-vandurme:2017:EACLshort}. For example, \Cref{fig:repr} illustrates a pair of input Chinese sentence and its English predicate-argument information\footnote{The predicate-argument information is usually denoted by relation tuples. In this work, we adopt the tree-structured representation generated by PredPatt~\cite{white-EtAl:2016:EMNLP2016}, which was a lightweight tool available at \url{https://github.com/hltcoe/PredPatt}.}, where predicate and argument are well used {\em semantic structure tags}.

\begin{figure}[t]
\centering
\includegraphics[width=0.49\textwidth]{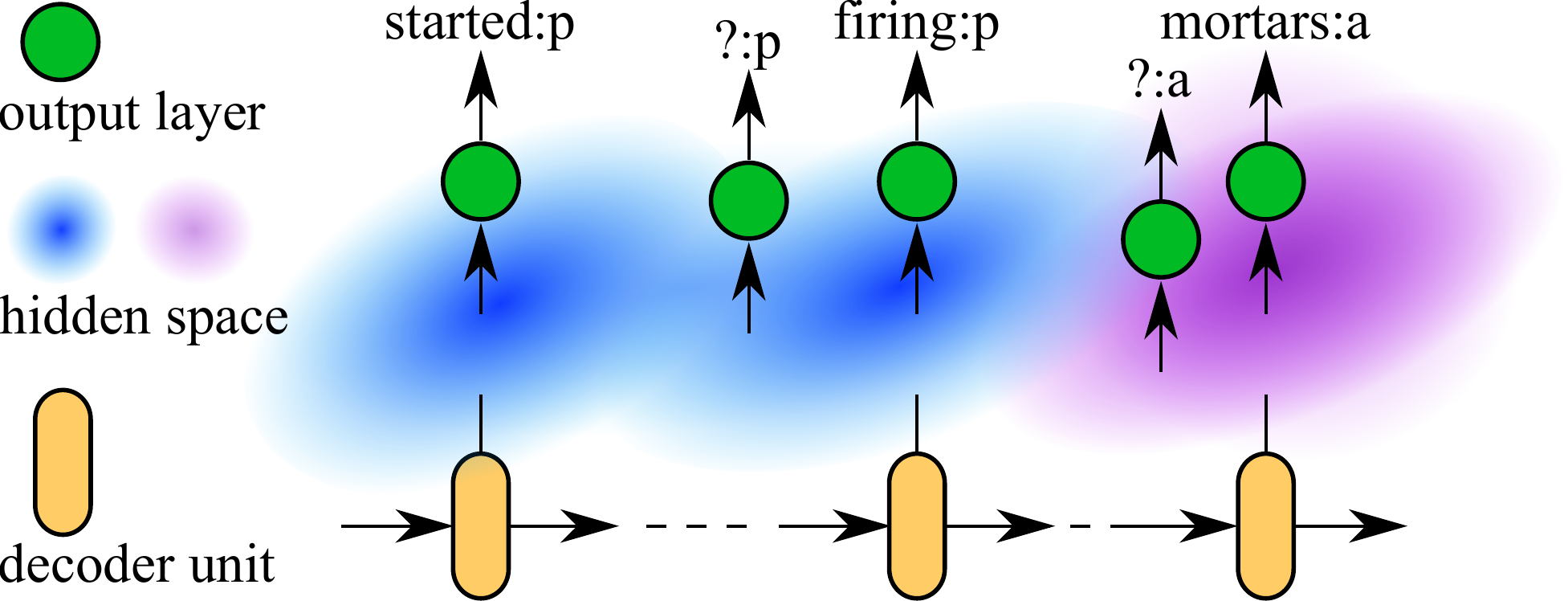}
\vspace{-17pt}
\caption{Visualization of {\em Halo} method. While a neural model learns to summarizes the current known information into a hidden state and predict the next target token, the surroundings of this hidden state in the same space (two-dimensional in this example) are supervised to generate tokens with the same semantic structure tag. For example, at the last shown step, the centroid of \textcolor{purple}{purple} area is the summarized hidden state and learns to predict `mortars:a', while a randomly sampled neighbor is enforced to generate an argument, although it may not be `mortars' (thus denoted by `?'). Similar remarks apply to the \textcolor{blue}{blue} regions.}
\vspace{-10pt}
\label{fig:halo}
\end{figure}

It is of great importance to solve the task, as to provide viable solutions to extracting information from the text of languages that suffer from no or little existing information extraction tools.
Neural models have empirically proven successful in this task~\cite{zhang-duh-vandurme:2017:EACLshort,zhang-EtAl:2017:IJCNLP}, but still remain unsatisfactory in low resource (i.e. small number of training samples) settings.
These neural models learn to summarize a given source sentence and target prefix into a hidden state, which aims to generate the correct next target token after being passed through an output layer. As each member in the target vocabulary is essentially either predicate or argument, a random perturbation on the hidden state should still be able to yield a token with the same semantic structure tag. This inductive bias motivates an extra term in training objective, as shown in \Cref{fig:halo}, which enforces the surroundings of any learned hidden state to generate tokens with the same semantic structure tag (either predicate or argument) as the centroid. We call this technique {\em Halo}, because the process of each hidden state taking up its surroundings is analogous to how the halo is formed around the sun. The method is believed to help the model generalize better, by learning more semantics-aware and noise-insensitive hidden states without introducing extra parameters.

\section{The Problem}
We are interested in learning a probabilistic model that directly maps an input sentence  $\{x_i\}_{i=1}^{I}= x_1 x_2 \ldots x_{I}$ of the source language $\mathcal{S}$ into an output sequence $\{y_t\}_{t=1}^{T}=y_1 y_2 \ldots y_{T}$ of the target language $\mathcal{T}$, where $\mathcal{S}$ can be any human natural language (e.g. Chinese) and $\mathcal{T}$ is the English PredPatt~\cite{white-EtAl:2016:EMNLP2016}. In the latter vocabulary, each type is tagged as either predicate or argument---those with ``:p'' are predicates while those with ``:a'' are arguments.

For any distribution $P$ in our proposed family, the {\em log-likelihood} $\ell$ of the model $P$ given any $(\{y_t\}_{t=1}^{T}\mid \{x_i\}_{i=1}^{I})$ pair is:
\begin{equation}\label{eqn:loglik-orig}
	\sum_{t=1}^{T} \log P\!\left( y_t \mid y_{t-1}, \ldots, y_{0}, \{x_i\}_{i=1}^{I}\right)
\end{equation}
where $y_0$ is a special beginning of sequence token.

We denote vectors by bold lowercase Roman letters such as $\vec{h}$, and matrices by bold capital Roman letters such as $\vec{W}$ throughout the paper.
Subscripted bold letters denote distinct vectors or matrices (e.g., $\vec{p}_t$).
Scalar quantities, including vector and matrix elements such as $h_d$ and $p_{t,y_t}$, are written without bold. Capitalized scalars represent upper limits on lowercase scalars, e.g., $1 \leq d \leq D$.
Function symbols are notated like their return type. All $\Real \rightarrow \Real$ functions are extended to apply elementwise to vectors and matrices.

\section{The Method}
In this section, we first briefly review how the baseline neural encoder-decoder models work on this task, and then introduce our novel and well-suited training method {\em Halo}.

\subsection{Baseline Neural Models}
Previous neural models on this task~\cite{zhang-duh-vandurme:2017:EACLshort,zhang-EtAl:2017:IJCNLP} all adopt an encoder-decoder architecture with recurrent neural networks, particularly LSTMs~\cite{hochreiter1997long}. At each step $t$ in decoding, the models summarize the input $\{x_i\}_{i=1}^{I}$ and output prefix $y_{1}, \ldots, y_{t-1}$ into a hidden state $\vec{h}_t \in (-1,1)^{D}$, and then project it with a transformation matrix $\vec{W} \in \Real^{|\mathcal{V}| \times D}$ to a distribution $\vec{p}_t$ over the target English PredPatt vocabulary $\mathcal{V}$:
\begin{subequations}\label{eqn:prob}
\begin{align}
\vec{p}_t &= {\vec{o}_t}/{(\vec{1}^\top \vec{o}_t})\\
\vec{o}_t &= \exp{\vec{W} \vec{h}_t} \in \Real_{+}^{|\mathcal{V}|}
\end{align}
\end{subequations}
where $\vec{1}$ is a $|\mathcal{V}|$-dimensional one vector such that $\vec{p}_t$ is a valid distribution.

Suppose that the ground truth target token at this step is $y_t$, the probability of generating $y_t$ under the current model is $p_{t, y_t}$, obtained by accessing the $y_t$-th element in the vector $\vec{p}_t$. Then the log-likelihood is constructed as $\ell = \sum_{t=1}^{T} \log p_{t,y_t}$, and the model is trained by maximizing this objective over all the training pairs.

\subsection{{\em Halo}}
Our method adopts a property of this task---the vocabulary $\mathcal{V}$ is partitioned into $\mathcal{P}$, set of predicates that end with ``:p'', and $\mathcal{A}$, set of arguments that end with ``:a''. As a neural model would summarize everything known up to step $t$ into $\vec{h}_t$, would a perturbation $\vec{h}'_t$ around $\vec{h}_t$ still generate the same token $y_t$? This bias seems too strong, but we can still reasonably assume that $\vec{h}'_t$ would generate a token with the same semantic structure tag (i.e. predicate or argument). That is, the prediction made by $\vec{h}'_t$ should end with ``:p'' if $y_t$ is a predicate, and with ``:a'' otherwise.

\begin{figure}[t]
\centering
\includegraphics[width=0.49\textwidth]{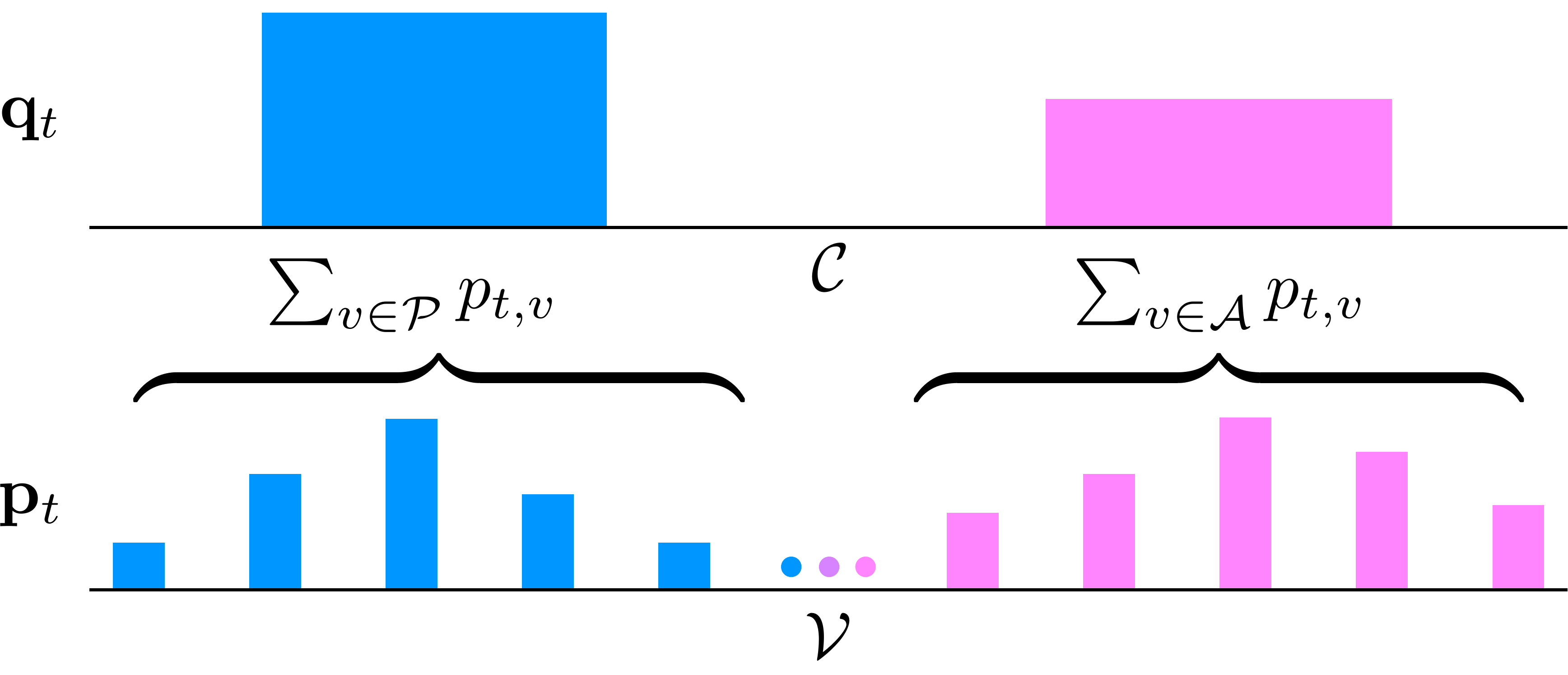}
\vspace{-17pt}
\caption{Visualization of how $\vec{q}$ (distribution over $\mathcal{C}$) is obtained by aggregating $\vec{p}$ (distribution over $\mathcal{V}$).}
\vspace{-10pt}
\label{fig:aggre}
\end{figure}

This inductive bias provides us with another level of supervision. Suppose that at step $t$, a neighboring $\vec{h}'_t$ is randomly sampled around $\vec{h}_t$, and is then used to generate a distribution $\vec{p}'_t$ in the same way as \cref{eqn:prob}. Then  we can get a distribution $\vec{q}'_t$ over $\mathcal{C}=\{ \text{predicate}, \text{argument} \}$, by summing all the probabilities of predicates and those of arguments:
\begin{subequations}\label{eqn:q}
\begin{align}
q'_{t,\text{predicate}} = \sum_{v \in \mathcal{P}} p'_{t,v}\\
q'_{t,\text{argument}} = \sum_{v \in \mathcal{A}} p'_{t,v}
\end{align}
\end{subequations}
This aggregation is shown in \Cref{fig:aggre}. Then the extra objective is $\ell' = \sum_{t=1}^{T} \log q'_{t,c_t}$, where $c_t=\text{predicate}$ if the target token $y_t \in \mathcal{P}$ (i.e. ending with ``:p'') and $c_t=\text{argument}$ otherwise.

Therefore, we get the joint objective to maximize by adding $\ell$ and $\ell'$:
\begin{equation}\label{eqn:loglikfinal}
	\ell + \ell' = \sum_{t=1}^{T} \log p_{t,y_t} + \sum_{t=1}^{T} \log q'_{t,c_t}
\end{equation}
which enables the model to learn more semantics-aware and noise-insensitive hidden states by enforcing the hidden states within a region to share the same semantic structure tag.\footnote{One can also sample multiple, rather than one, neighbors for one hidden state and then average their $\log q'_{t, c_t}$. In our experimental study, we only try one for computational cost and found it effective enough.}

\subsubsection{Sampling Neighbors}
Sampling a neighbor around $\vec{h}_t$ is essentially equivalent to adding noise to it.
Note that in a LSTM decoder that previous work used, $\vec{h}_t \in (-1,1)^D$ because $\vec{h}_t = \vec{o}_t \odot \tanh (\vec{c}_t)$ where $\vec{o}_t \in (0,1)^{D}$ and $\tanh (\vec{c}_t) \in (-1, 1)^D$. Therefore, extra work is needed to ensure $\vec{h}'_t \in (-1,1)^{D}$.
For this purpose, we follow the recipe\footnote{Alternatives do exist. For example, one can transform $\vec{h}_t$ from $(-1, 1)^{D}$ to $(-\infty, \infty)^{D}$, add random (e.g. Gaussian) noise in the latter space and then transform back to $(-1, 1)^{D}$. These tricks are valid as long as they find neighbors within the same space $(-1, 1)^{D}$ as $\vec{h}_t$ is.}:
\begin{itemize}[noitemsep]
\item Sample $\vec{h}''_t \in (-1,1)^D$ by independently sampling each entry from an uniform distribution over $(-1,1)$;
\item Sample a scalar $\lambda_t \in (0,1)$ from a Beta distribution $B(\alpha, \beta)$ where $\alpha$ and $\beta$ are hyperparameters to be tuned;
\item Compute $\vec{h}'_t = \vec{h}_t + \lambda_t (\vec{h}''_t - \vec{h}_t)$ such that $\vec{h}'_t \in (-1,1)^D$ lies on the line segment between $\vec{h}_t$ and $\vec{h}''_t$.
\end{itemize}

Note that the sampled hidden state $\vec{h}'_t$ is only used to compute $\vec{q}'_t$, but not to update the LSTM hidden state, i.e., $\vec{h}_{t+1}$ is independent of $\vec{h}'_t$.

\subsubsection{Roles of Hyperparameters}\label{sec:hype}
The {\em Halo} technique adds an inductive bias into the model, and its magnitude is controlled by $\lambda_t$:
\begin{itemize}[noitemsep]
\item $\lambda_t \in (0,1)$ to ensure $\vec{h}'_t \in (-1,1)^D$;
\item $\lambda_t \rightarrow 0$ makes $\vec{h}'_t \rightarrow \vec{h}_t$, thus providing no extra supervision on the model;
\item $\lambda_t \rightarrow 1$ makes $\vec{h}'_t$ uniformly sampled in entire $(-1,1)^D$, and causes underfitting just like a $L$-2 regularization coefficient goes to infinity.
\end{itemize}

We sample a valid $\lambda_t$ from a Beta distribution with $\alpha > 0$ and $\beta > 0$, and their magnitude can be tuned on the development set:
\begin{itemize}[noitemsep]
\item When $\alpha \rightarrow 0$ and $\beta$ is finite, or $\alpha$ is finite and $\beta \rightarrow \infty$, we have $\lambda_t \rightarrow 0$;
\item When $\alpha \rightarrow \infty$ and $\beta$ is finite, or $\alpha$ is finite and $\beta \rightarrow 0$, we have $\lambda_t \rightarrow 1$;
\item Larger $\alpha$ and $\beta$ yield larger variance of $\lambda_t$, and setting $\lambda_t$ to be a constant is a special case that $\alpha \rightarrow \infty$, $\beta \rightarrow \infty$ and $\alpha/\beta$ is fixed.
\end{itemize}

Besides $\alpha$ and $\beta$, the way of partitioning $\mathcal{V}$ (i.e. the definition of $\mathcal{C}$) also serves as a knob for tuning the bias strength. Although on this task, the predicate and argument tags naturally partition the vocabulary, we are still able to explore other possibilities. For example, an extreme is to partition $\mathcal{V}$ into $|\mathcal{V}|$ different singletons, meaning that $\mathcal{C}=\mathcal{V}$---a perturbation around $\vec{h}_t$ should still predict the same token. But this extreme case does not work well in our experiments, verifying the importance of the semantic structure tags on this task.

\begin{table*}[t]
\begin{center}
\begin{small}
\begin{sc}
\begin{tabularx}{0.99\textwidth}{l *{3}{R}*{2}{R}*{1}{R}}
\toprule
Dataset & \multicolumn{3}{c}{Number of Pairs} & \multicolumn{2}{c}{Vocabulary Size} & {Token/Type} \\
\cmidrule(lr){2-4} \cmidrule(lr){5-6}
  & Train & Dev & Test & Source & Target & \\
\midrule
{\sc Chinese} & $941040$ & $10000$ & $39626$ & $258364$ & $234832$ & $91.94$ \\
{\sc Uzbek} & $31581$ & $1373$ & $1373$ & $69255$ & $37914$ & $12.18$ \\
{\sc Turkish} & $20774$ & $903$ & $903$ & $51248$ & $32009$ & $11.97$ \\
{\sc Somali} & $10702$ & $465$ & $465$ & $29591$ & $18616$ & $12.78$ \\
\bottomrule
\end{tabularx}
\end{sc}
\end{small}
\end{center}
\vspace{-10pt}
\caption{Statistics of each dataset.}
\label{tab:data}
\vspace{-0pt}
\end{table*}

\begin{table*}[t]
\begin{center}
\begin{small}
\begin{sc}
\begin{tabularx}{1.0\linewidth}{l *{3}{R}*{3}{R}*{3}{R}*{3}{R}}
\toprule
Method & \multicolumn{3}{c}{Chinese} & \multicolumn{3}{c}{Uzbek} & \multicolumn{3}{c}{Turkish} & \multicolumn{3}{c}{Somali}\\
\cmidrule(lr){2-4} \cmidrule(lr){5-7} \cmidrule(lr){8-10} \cmidrule(lr){11-13}
  & BLEU & \multicolumn{2}{c}{F1} & BLEU & \multicolumn{2}{c}{F1} & BLEU & \multicolumn{2}{c}{F1} & BLEU & \multicolumn{2}{c}{F1} \\
  &  & Pred & Arg &  & Pred & Arg &  & Pred & Arg &  & Pred & Arg \\
\cmidrule(lr){2-4} \cmidrule(lr){5-7} \cmidrule(lr){8-10} \cmidrule(lr){11-13}
ModelZ & 22.07 & 30.06 & 39.06 & 10.76 & 12.46 & 24.08 & 7.47 & 6.49 & 17.76 & 13.06 & 13.91 & 25.38 \\
ModelP & 22.10 & 30.04 & 39.83 & 12.50 & 18.81 & 25.93 & 9.04 & \textbf{12.90} & 21.13 & 13.22 & 16.71 & 26.83 \\
ModelP-{\em Halo} & \textbf{23.18} & \textbf{30.85} & \textbf{41.23} & \textbf{12.95} & \textbf{19.23} & \textbf{27.63} & \textbf{10.21} & 12.55 & \textbf{22.57} & \textbf{14.26} & \textbf{17.06} & \textbf{27.73} \\
\bottomrule
\end{tabularx}
\end{sc}
\end{small}
\end{center}
\vspace{-10pt}
\caption{{\sc Bleu} and {\sc F1} scores of different models on all these datasets, where {\sc Pred} stands for predicate and {\sc Arg} for argument. Best numbers are highlighted as \textbf{bold}.}
\label{tab:bleu}
\vspace{-10pt}
\end{table*}

\section{Related Work}
Cross-lingual information extraction has drawn a great deal of attention from researchers. Some~\cite{sudo2004cross,parton09crosslingual,ji09,snover11slot,ji16tac} worked in closed domains, i.e. on a predefined set of events and/or entities, \citet{zhang-duh-vandurme:2017:EACLshort} explored this problem in open domain and their attentional encoder-decoder model significantly outperformed a baseline system that does translation and parsing in a pipeline.
\citet{zhang-EtAl:2017:IJCNLP} further improved the results by inventing a hierarchical architecture that learns to first predict the next semantic structure tag and then select a tag-dependent decoder for token generation. Orthogonal to these efforts, {\em Halo} aims to help {\em all neural models on this task}, rather than any specific model architecture.

{\em Halo} can be understood as a data augmentation technique~\cite{chapelle2001,vandermaaten2013corrupted,srivastava2014dropout,szegedy2016rethinking,gal2016theoretically}.
Such tricks have been used in training neural networks to achieve better generalization, in applications like image classification~\cite{simard2000,simonyan2015very,arpit2017closer,zhang2017mixup} and speech recognition~\cite{graves2013speech,amodei2016deep}.
{\em Halo} differs from these methods because 1) it makes use of the task-specific information---vocabulary is partitioned by semantic structure tags; and 2) it makes use of the human belief that the hidden representations of tokens with the same semantic structure tag should stay close to each other.
Some

\section{Experiments}\label{sec:exp}
We evaluate our method on several real-world CLIE datasets measured by {\sc BLEU}~\cite{papineni2002bleu} and {\sc F1}, as proposed by~\citet{zhang-duh-vandurme:2017:EACLshort}. For the generated linearized PredPatt outputs and their references, the former metric\footnote{The MOSES implementation~\cite{koehn2007moses} was used as in all the previous work on this task.} measures their n-gram similarity, and the latter measures their token-level overlap. In fact, {\sc F1} is computed separately for predicate and argument, as {\sc F1 Pred} and {\sc F1 Arg} respectively.

\subsection{Datasets}
\label{sec:data}
Multiple datasets were used to demonstrate the effectiveness of our proposed method, where one sample in each dataset is a source language sentence paired with its linearized English PredPatt output.
These datasets were first introduced as the DARPA LORELEI Language Packs~\cite{strassel2016lorelei}, and then used for this task by \citet{zhang-duh-vandurme:2017:EACLshort,zhang-EtAl:2017:IJCNLP}.
As shown in \cref{tab:data}, the {\sc Chinese} dataset has almost one million training samples and a high token/type ratio, while the others are {\em low resourced}, meaning they have much fewer samples and lower token/type ratios.

\subsection{Model Implementation}
Before applying our {\em Halo} technique, we first improved the current state-of-the-art neural model of \citet{zhang-EtAl:2017:IJCNLP} by using residual connections \cite{he2016residual} and multiplicative attention \cite{luong2015effective}, which effectively improved the model performance. We refer to the model of \citet{zhang-EtAl:2017:IJCNLP} and our improved version as ModelZ and ModelP respectively\footnote{Z stands for Zhang and P for Plus.}.

\subsection{Experimental Details}
In experiments, instead of using the full vocabularies shown in \cref{tab:data}, we set a minimum count threshold for each dataset, to replace the rare words by a special out-of-vocabulary symbol. These thresholds were tuned on dev sets.

The Beta distribution is very flexible. In general, its variance is a decreasing function of $\alpha+\beta$, and when $\alpha+\beta$ is fixed, the mean is an increasing function of $\alpha$. In our experiments, we fixed $\alpha+\beta=20$ and only lightly tuned $\alpha$ on dev sets. Optimal values of $\alpha$ stay close to $1$.

\subsection{Results}
As shown in \Cref{tab:bleu}, ModelP outperforms ModelZ on all the datasets measured by all the metrics, except for {\sc F1 Pred} on {\sc Chinese} dataset. Our {\em Halo} technique consistently boosts the model performance of {\sc ModelP} except for {\sc F1 Pred} on {\sc Turkish}.

Additionally, experiments were also conducted on two other low resource datasets {\sc Amharic} and {\sc Yoruba} that \citet{zhang-EtAl:2017:IJCNLP} included, and $\alpha=0$ in {\em Halo} was found optimal on the dev sets. In such cases, this regularization was not helpful so no comparison need be made on the held-out test sets.

\section{Conclusion and Future Work}
We present a simple and effective training technique {\em Halo} for the task of cross-lingual information extraction. Our method aims to enforce the local surroundings of each hidden state of a neural model to only generate tokens with the same semantic structure tag, thus enabling the learned hidden states to be more aware of semantics and robust to random noise. Our method provides new state-of-the-art results on several benchmark cross-lingual information extraction datasets, including both high and low resource scenarios.

As future work, we plan to extend this technique to similar tasks such as POS tagging and Semantic Role Labeling. One straightforward way of working on these tasks is to define the vocabularies as set of `word-type:POS-tag' (so $c_t=\text{POS tag}$) and `word-type:SR' (so $c_t=\text{semantic role}$), such that our method is directly applicable. It would also be interesting to apply {\em Halo} widely to other tasks as a general regularization technique.

\section*{Acknowledgments}
This work was supported in part by the JHU Human Language Technology Center of Excellence (HLTCOE), and DARPA LORELEI. The U.S. Government is authorized to reproduce and distribute reprints for Governmental purposes. The views and conclusions contained in this publication are those of the authors and should not be interpreted as representing official policies or endorsements of DARPA or the U.S. Government.
\bibliography{references}
\bibliographystyle{acl_natbib}

\end{document}